\documentclass{article} 
\usepackage{nips12submit_e,times}
\usepackage{hyperref}
\usepackage{graphicx} 
\usepackage{tabularx}
\usepackage{natbib}
\usepackage{verbatim}

\usepackage{my_gp_macros}
\usepackage{amsmath}
\usepackage{amssymb}
\usepackage{amsfonts}
\usepackage{parskip}
\usepackage{paralist}

\usepackage{caption}
\usepackage{subcaption}
\newcommand{\best}[1]{\ensuremath{\boldsymbol{#1}}}

\newcommand{\inlinepara}[1]{\textbf{#1}~~}

\title{Transductive Log Opinion Pool \\
  of Gaussian Process Experts}
\author{
	Yanshuai Cao ~~
	David J.~Fleet \\
	Department of Computer Science, ~~
	University of Toronto
}

\newcommand{\Tau}{\mathrm{T}}

\newcommand{\colem}[1]{\textcolor{blue}{#1}}

\nipsfinalcopy 

\begin{document}
\maketitle

\vspace*{-0.4cm}
\begin{abstract}
We introduce a framework for analyzing transductive combination of Gaussian process (GP) experts, where independently trained GP experts are combined in a way that depends on test point location, in order to scale GPs to big data. The framework provides some theoretical justification for the generalized product of GP experts (gPoE-GP) which was previously shown to work well in practice \cite{gpoe,rbcm} but lacks theoretical basis. Based on the proposed framework, an improvement over gPoE-GP is introduced and empirically validated.
\end{abstract}

\vspace*{-0.5cm}
\section{Introduction}
\vspace*{-0.2cm}

Gaussian processes (GP) are nonparametric Bayesian models that give probabilistic predictions required in many tasks where uncertainty is important. However, training and inference scale cubically and quadratically in the number of training data. In order to scale GP for Big Data, a number of approaches have been proposed. Most methods assume some form of conditional independence given some ``inducing points'', yielding sparse GPs \cite{Snelson06,titsias,cholqr,hensman}. While sparse GPs are much faster than the full GP, the hyperparameter optimization in these sparse GPs are still very slow in practice. 

An alternative approach is to learn separate GPs on subsets of training data, and combine their predictions at test time, e.g.\ the Bayesian committee machine\cite{bcm}. We refer to this class of methods as transductive combination of GP experts. This approach has low computational cost and is easy to parallelize. In recent work \cite{gpoe,rbcm}, a new heuristic that reweights expert predictions, based on the change in entropy of expert GPs at test points, allows transductive combination of GP experts to achieve state-of-art results among large scale GP methods on several regression datasets. Nevertheless, there has been little theoretical justification for this new idea. In this work, we introduce a framework for transductive combination of GP experts that explains recent empirical successes. The framework also suggests further improvement, which we validate empirically. The improved model, called diversified log opinion pool of GP experts (dLOP-GP), is highly scalable, has a sound theoretical basis, and also improves over the previous state-of-art results.

\vspace*{-0.3cm}
\section{Transductive combination of GP experts}
\vspace*{-0.3cm}

The idea behind transductive combination of GPs is to train separate GP experts on subsets of data independently. Then, at test time, expert predictions are combined in a way that depends on the test point location. This scheme nicely balances the training and testing computational cost with overall model expressiveness. Learning is data-parallelizable, while for prediction, the computation nicely follows the map-reduce paradigm. Each expert computes its prediction independently during the map phase and their predictions are combined together in the reduce phase. Compared to local transductive GP \cite{local_gp_reg}, where local GPs are trained on subsets of training points selected based on test locations, transductive combination of GP experts requires no GP training at test time, hence it scales better for prediction. Furthermore, because the way expert predictions are combined depends on test location, rather than being fixed, the resulting model is much more expressive than simply taking an average of predictions. This approach can also potentially model heteroscedasticity and nonstationarity, even though individual expert GPs use relatively simple stationary kernels.


\inlinepara{BCM}
The Bayesian committee machine \cite{bcm} is the first well known transductive combination of GPs. Given input-output pairs, $\{(x_i,f_i)\}_i$, a GP prior $p(f)$, and a partition of training data $D$ into disjoint subsets $\{D_i\}$, i.e. $D_j \cap D_i = \emptyset$. If we assume that when conditioned on a test point, $(x_\star, f_\star)$, the subsets of data are independent, then the posterior distribution at the test point is: 
\begin{equation}
p(f_\star | x_\star, D) = \frac{1}{Z}\frac{\prod^{K}_{i}p(f_\star | x_\star, D_i)}{p^{K-1}(f_\star | x_\star)}
\end{equation}
The resulting predictive mean and variance are: $m_{bcm}(x_\star) = \left(\sum_i m_i(x_\star) \Tau_i(x_\star)\right)\sigma^2_{bcm}(x_\star)$ and $\sigma^2_{bcm}(x_\star) = {\left(\sum_i \Tau_i(x_\star) + (1 -M)\Tau_{\star\star} \right)}^{-1}$, where $m_i(x_\star)$ and $\Tau_i(x_\star) = \sigma_i^{-2}(x_\star)$ are the mean and the precision of the $i$-th Gaussian expert at point $x_\star$ respectively, and $\Tau_{\star\star}$ is the prior precision. Because BCM is derived from the conditional independence assumption and a common prior $p(f)$, using Bayes' rule, the GP on each subset of data $D_i$ has to share the same kernel and kernel hyperparameters. Therefore, it does not naturally provide a model where different regions of space are modelled using different kernels or hyperparameters. Beyond this lack of expressiveness, empirically, BCM has been shown to under-perform by a large margin\cite{rbcm}.

\inlinepara{gPoE-GP}
The generalized product of GP experts \cite{gpoe} combines experts as follows:
\begin{equation}
	p(f_\star|x_\star, D) = \frac{1}{Z}\prod_{i}p_i^{\alpha_i(x)}(f_\star|x_\star, D_i) \label{eq:gpoe}
\end{equation}
where $\alpha_i(x_\star) \ge 0$ and $\sum_i \alpha_i(x_\star) = 1$. Here, $\alpha_i(x_\star)$ is a measure of reliability, which re-weights each expert's prediction. 
Taking $\alpha_i(x_\star)$ to be proportional to change between prior and posterior entropy of the Gaussian distribution of the $i$th expert at point $x_\star$, is simple and effective in practice \cite{gpoe}.

The predictive mean and variance are: $m_{gpoe}(x_\star) = (\sum_i m_i(x_\star) \alpha_i(x_\star)\Tau_i(x_\star) )\sigma^2_{gpoe}(x_\star)$, $\sigma^2_{gpoe}(x_\star) = {(\sum_i \alpha_i(x_\star)\Tau_i(x_\star))}^{-1}$. With the constraint that $\sum_i \alpha_i=1$, the predictive distribution falls back to prior when far from training data. Nevertheless, the overall model is ad-hoc without a solid foundation, and was previously justified solely by its empirical performance \cite{gpoe,rbcm}.


\inlinepara{rBCM}
The robust Bayesian committee machine \cite{rbcm} inherits the theoretical basis of the BCM, and it incorporates the reweighting by the change in entropy, like the gPoE-GP (the $\alpha_i(x_\star)$). However, the BCM framework does not provide an explanation for the reweighting factor, and the rBCM is limited to all the restrictions of BCM, that is, subsets of data for the different experts need to be disjoint, and experts need to share the same kernel specification and hyperparameters. 

\vspace*{-0.2cm}
\section{Transductive log opinion pool framework}
\vspace*{-0.2cm}

In what follows we formulate the log opinion pool of GP experts. The model is strongly motivated by the log opinion pool in \cite{lop}, but adjusted for the transductive case. 


Assume there are $K$ GP experts $\{p_i\}^K_{i=1}$, each of which gives a posterior distribution $p_i(f_\star|x_\star, D_i)$ at the test point $x_\star$. Also, assume we have some measure of the relative reliability of each expert at the test point, denoted by $\alpha_i(x_\star)$, satisfying $\sum_i \alpha_i(x_\star)=1$ and $\alpha_i(x_\star) \ge 0$. Then, we want to find a distribution $\widetilde{p}(f_\star | x_\star)$ that is close to all experts in a weighted KL sense:
\begin{equation}
\widetilde{p}(f_\star|x_\star) = \arg\!\!\min_{p(f_\star|x_\star)}{\sum_{i}\alpha_i KL({p\|p_i}})
\end{equation}
where the only unknowns are the $\alpha_i(x_\star)$'s. Under the constraint that $\int p(f_\star|x_\star) d f_\star=1$, the solution is exactly the gPoE in Eq. \ref{eq:gpoe} \cite{lop}. Hence, this new framework provides a sound theoretical basis for gPoE, i.e,\ it results from model averaging in the sense of weighted KL divergence. This new interpretation has two important implications: first, no form of conditional independence assumption is required anymore; second, the expert GPs do not need to share a common prior. In other words, the subsets $\{D_i\}_i$ need not be disjoint, and expert GPs can have different hyperparameters or even different kernel specification.

In \cite{lop}, $\alpha_i$'s are fixed for each expert, whereas here they vary with location $x_\star$. Another major difference is that in \cite{lop}, the unknown $p_\star$ can be replaced by the training inputs-outputs pairs (empirical distribution) so that $\alpha_i$'s can be learned via optimization. But in the transductive setting, outputs are not observed, so inference for $\alpha_i(x_\star)$ needs to be done differently.

\vspace*{-0.2cm}
\section{Selecting weights in transductive log opinion pool of GPs}
\vspace*{-0.2cm}
We now explore alternative ways to infer the weights $\alpha_i(x_\star)$. For notational convenience, we drop the argument and write $\alpha_i$ directly henceforth.
We need to analyze the effect of $\alpha_i$ on the KL divergence of $\widetilde{p}(f_\star|x_\star)$ from the unknown ground-truth distribution $p_{\star}(f_\star|x_{\star})$. In \cite{lop}, it was proved that: 
\begin{equation}
  KL(p_\star \| \widetilde{p}) = \sum_i\alpha_i KL({p_\star\|p_i}) - \sum_i\alpha_i KL({\widetilde{p} \| p_i}) \label{eq:true_obj}
\end{equation}
Because $\widetilde{p}$ implicitly depends on $\alpha_i$'s as well, it is not clear from Eq.\ref{eq:true_obj} what the best way is to select $\alpha_i$'s. However, as shown in \cite{lop}, $KL(p_\star \| \widetilde{p})$ can be well approximated by:
\begin{equation}
  KL(p_\star \| \widetilde{p}) \approx \sum_i\alpha_i KL({p_\star\|p_i}) - \frac{1}{4}\sum_{i,j}\alpha_i \alpha_j (KL({p_i \| p_j}) + KL({p_j \| p_i})) \equiv E - C \label{eq:approx_obj}
\end{equation}
where $E$ and $C$ are defined to be the first and second terms in the approximation respectively.
Eq.\ref{eq:approx_obj} still cannot be used as an objective to optimize for $\alpha_i$'s numerically, because the unknown $p_\star$ cannot be approximated by samples in the transductive setting. However, Eq.\ref{eq:approx_obj} does provide useful insight:

\inlinepara{Explaining the entropy change heuristic of gPoE-GP}
To reduce $KL(p_\star \| \widetilde{p})$, one needs to decrease $E$ and increase $C$. Decreasing $E$ entails setting higher weights $\alpha_i$ for experts that predict well at the test point, i.e. small $KL(p_\star \| p_i)$, and vice versa. Again because $p_\star$ is unknown, one cannot actually compute the terms $KL(p_\star \| p_i)$. But we know that, without seeing any training data, the prior distribution is unlikely to be close to $p_\star$ on average. So we can decrease the weights of the experts whose prediction are not significantly influenced by training data. With Gaussian process experts, the change in entropy between prior and posterior at point $x_i$ provides such a measure. gPoE-GP ignores the second term $C$ in Eq.\ref{eq:approx_obj}, and attempts to decrease the first term by lowering relative weights of experts whose predictions are less influenced by data. When benefits of diversification are small, e.g.\ if subsets of points are uniformly randomly sampled and all experts share kernel hyper-parameters, ignoring $C$ could be a reasonable approximation. 

\inlinepara{Diversified log opinion pool of GP experts}
Increasing the second term $C$ in Eq.\ref{eq:approx_obj} entails increasing weights for pairs of predictions that are further apart in the sense of symmetric KL divergence. This encourages more diversified predictions. Because $C$ does not include $p_\star$, it can be computed. In fact, $C$ is a quadratic form in the weight (row) vector $\boldsymbol{\alpha}=[\alpha_i]^K_{i=1}$: 
$C = \frac{1}{4}\boldsymbol{\alpha}Q\boldsymbol{\alpha}^{\T}$, 
where, $Q$ is a $K$x$K$ symmetric matrix whose $ij$-entry is $Q_{ij}=KL({p_i \| p_j}) + KL({p_j \| p_i})$.


A simple way to incorporate $C$ into the determination of $\{\alpha_i\}^K_{i=1}$ is to start with the $\alpha_i$'s of gPoE-GP as initialization, and modify them in a way that increases $C$. Of course, this does not always guarantee a decrease in Eq.\ref{eq:approx_obj}, but we find empirically that simply taking a single normalized gradient step of the $C$ term then re-normalizing $\sum_i \alpha_i$ to $1$ works well. Since Eq.\ \ref{eq:approx_obj} is already an approximate objective, and we cannot truly optimize it due to the $E$ term, minimizing the $C$ term more accurately than this may not be worth the effort. Because the resulting model encourages more diverse expert predictions than gPoE-GP, we refer to it as diversified log opinion pool of GP experts (dLOG-GP). In other words, at a test point $x_\star$, the weight vector $\boldsymbol{\alpha}(x_\star)$ for experts in dLOP-GP is computed as following: first, set $\boldsymbol{\alpha}=\boldsymbol{\alpha^{gpoe}}$, where $\boldsymbol{\alpha^{gpoe}}$ is set to the changes in entropy of the experts, and re-normalized to sum to $1$; then set $\boldsymbol{\hat{\alpha}} = \boldsymbol{\alpha^{gpoe}} + \lambda \frac{\bigtriangledown C}{\|\bigtriangledown C\|}$; and finally, $\boldsymbol{\alpha} = {\boldsymbol{\hat{\alpha}}}/{\sum_i{\hat{\alpha}_i}}$ where $\hat{\alpha}_i=\boldsymbol{\hat{\alpha}}[i]$. We found that a range of step sizes, $\lambda$, all work well in practice, as well as taking more than one gradient step, so we simply use $\lambda = 1.0$ and use one update step in the experiments below. The forms of predictive mean and variance are the same as gPoE-GP.

\vspace*{-0.3cm}
\section{Experiment}
\vspace*{-0.2cm}
Next we report empirical results on the dLOP-GP along with several related models. To this end we use three standard regression datasets: KIN40K (8D feature space, 10K training points, 30K test points), SARCOS (21D, $44,484$ training points, $4,449$ test points), and the UK apartment price dataset (2D, $64,910$ training points, 10K test points) used in \cite{hensman}. 

For each model on each dataset, we explore five different ways to select subsets of data used by experts, and show that dLOP-GP generally improves over previous methods. The five subset selection approaches are:
\begin{inparaenum}[\itshape i\upshape)]
\item ({\em \bf DS}) points are randomly partitioned into disjoint subsets, experts share the same hyperparameters learned from a random subset of data GP; 
\item ({\em \bf SoD-Shared-Hyp}) subsets of data are randomly sampled with replacement (so not necessarily disjoint), experts share the same hyperparameters as in DS;
\item ({\em \bf SoD}) same as the previous SoD-Shared-Hyp scheme but experts have different hyperparameters that are independently learned;
\item ({\em \bf tree}) subsets are based on a tree partition: a ball tree \cite{balltree} built on training set recursively partitions the space; and on each level of the tree, a random subset of data is drawn to build a GP; kernel hyperparameters are independently learned;
\item ({\em \bf tree-rand-kern}) same as the previous tree based construction, but the kernels for experts are randomly specified as square exponential ARD kernel, Matern32, Matern52 or sum of any of these three.
\end{inparaenum} 
gPoE-GP and dLOP-GP both work with all construction schemes. For rBCM, only the DS construction is valid in theory, as the sets are disjoint with the same hyperparameters, but we are not able to replicate results from \cite{rbcm}, especially when predictive uncertainty is taken into account in benchmarking. rBCM does not truly support the other schemes, since experts do not share the same prior anymore, but for comparison, we run it by taking the average of all experts' prior variance, instead of a common prior variance in these other scenarios. 

All subsets have $512$ points (except for one of the subsets in the DS scheme, whenever the total number of data is not divisible by $512$). And with the exception of $DS$, all schemes use $128$ experts. The only overhead that dLOP-GP adds compared to gPoE-GP is in the one gradient step to determined $\alpha_i$'s. Empirically, this presents only a $5\%$ to $10\%$ increases in test computation time. In our experiments, on KIN40K under the SoD scheme, training takes about $20$ minutes when parallelized on 24 cores, while testing takes about $400$ seconds for gPoE-GP and $420$ seconds for dLOP-GP. 
\vspace*{-0.2cm}
\begin{table}[h]
\resizebox{.74\textwidth}{!}{%
\begin{subtable}[t]{\linewidth}
\begin{tabular}{|c|c|c|c|}
\hline
~ & dLOP & gPoE &  rBCM \\ \hline
DS & $(\best{-0.542 \pm 0.098}, \best{0.360 \pm 0.066})$ & $(-0.541 \pm 0.108, 0.361 \pm 0.073)$ & $(1.507 \pm 2.244, 0.405 \pm 0.071)$ \\ \hline
Sod-Shared-Hyp & $(\best{-0.589 \pm 0.194}, \best{0.287 \pm 0.108})$ & $(-0.570 \pm 0.212, 0.316 \pm 0.138)$ & $(5.233 \pm 5.079, 0.379 \pm 0.167)$ \\ \hline
SoD & $(\colem{\best{-0.922 \pm 0.057}, \best{0.161 \pm 0.023}})$ & $(-0.833 \pm 0.047, 0.187 \pm 0.021)$ & $(1.537 \pm 0.603, 0.184 \pm 0.021)$ \\ \hline
Tree & $(\best{-0.767 \pm 0.066}, 0.231 \pm 0.036)$ & $(-0.765 \pm 0.014, 0.214 \pm 0.013)$ & $(1.014 \pm 0.703, \best{0.192 \pm 0.015})$ \\ \hline
Tree-Rand-Kern & $(\best{-0.894 \pm 0.054}, \best{0.179 \pm 0.028})$ & $(-0.818 \pm 0.037, 0.205 \pm 0.017)$ & $(0.195 \pm 0.333, 0.312 \pm 0.240)$ \\
\hline
\end{tabular}
\caption{KIN40K} \label{t:kin40k}
\end{subtable}
}
\resizebox{.717\textwidth}{!}{%
\begin{subtable}[t]{\linewidth}
\begin{tabular}{|c|c|c|c|}
\hline
~ & dLOP & gPoE &  rBCM \\ \hline
DS & $(0.639 \pm 4.006, \best{0.253 \pm 0.372})$ & $(\best{0.562 \pm 4.046}, 0.259 \pm 0.369)$ & $(5.390 \pm 7.706, 0.271 \pm 0.383)$ \\ \hline
Sod-Shared-Hyp & $(\best{-0.287 \pm 0.085}, \best{0.489 \pm 0.091})$ & $(-0.276 \pm 0.081, 0.499 \pm 0.089)$ & $(-0.094 \pm 0.064, 0.849 \pm 0.094)$ \\ \hline
SoD & $(-1.577 \pm 0.210, 0.046 \pm 0.006)$ & $(\best{-1.669 \pm 0.085}, 0.050 \pm 0.008)$ & $(4.505 \pm 3.629, \best{0.041 \pm 0.005})$ \\ \hline
Tree & $(\best{-2.164 \pm 0.039}, 0.029 \pm 0.003)$ & $(-1.999 \pm 0.054, 0.042 \pm 0.007)$ & $(-0.116 \pm 0.910, \best{0.028 \pm 0.003})$ \\ \hline
Tree-Rand-Kern & $(\colem{\best{-2.612 \pm 0.236}}, 0.022 \pm 0.004)$ & $(-2.507 \pm 0.193, 0.031 \pm 0.005)$ & $(-0.672 \pm 1.103, \colem{\best{0.020 \pm 0.004}})$ \\
\hline
\end{tabular}
\caption{SARCOS} \label{t:sarcos}
\end{subtable}
}
\resizebox{.73\textwidth}{!}{%
\begin{subtable}[t]{\linewidth}
\begin{tabular}{|c|c|c|c|}
\hline
~ & dLOP & gPoE &  rBCM \\ \hline
DS & $(\best{-0.200 \pm 0.012}, 0.002 \pm 0.000)$ & $(-0.199 \pm 0.011, 0.002 \pm 0.000)$ & $(1367.626 \pm 186.653, 4.352 \pm 0.438)$ \\ \hline
Sod-Shared-Hyp & $(\best{-0.208 \pm 0.006}, 0.002 \pm 0.000)$ & $(-0.207 \pm 0.006, 0.002 \pm 0.000)$ & $(1255.057 \pm 167.159, 4.085 \pm 0.329)$ \\ \hline
SoD & $(\best{-0.213 \pm 0.010}, 0.002 \pm 0.000)$ & $(-0.208 \pm 0.005, 0.002 \pm 0.000)$ & $(1350.815 \pm 39.045, 4.212 \pm 0.097)$ \\ \hline
Tree & $(\best{-0.375 \pm 0.007}, 0.002 \pm 0.000)$ & $(-0.336 \pm 0.008, 0.002 \pm 0.000)$ & $(181.482 \pm 2.472, 0.542 \pm 0.007)$ \\ \hline
Tree-Rand-Kern & $(\colem{\best{-0.379 \pm 0.004}}, 0.002 \pm 0.000)$ & $(-0.340 \pm 0.004, 0.002 \pm 0.000)$ & $(148.745 \pm 2.567, 0.483 \pm 0.008)$ \\
\hline
\end{tabular}
\caption{UK-APT} \label{t:ukapt}
\end{subtable}
}
\caption{Results across three models and five different subset selection methods on three regression datasets. Each tuple of result is (SNLP, SMSE). Best across the three models on each line is in bold, and best on each dataset is also coloured blue. In case of ties, nothing is marked.} 
\end{table}
\vspace*{-0.2cm}

Results are measured by standardized negative log probability (SNLP) and standardized mean square error (SMSE) in Table.\ref{t:kin40k} - \ref{t:ukapt}. SNLP is a more informative metric as it takes into account the uncertainty in prediction. Insights from the theoretical frameworks are confirmed: dLOP-GP almost always improves over gPoE-GP; and as individual experts become more diverse from SoD-Shared-Hyp to SoD, both gPoE-GP and dLOP-GP improve; and more variabilities among experts could lead to further improvement, as shown in the case of Tree and Tree-Rand-Kern on SARCOS and UK-APT datasets. Finally, the results also demonstrate the need for different rather than shared hyperparameters; only two models under the transductive log opinion pool framework, i.e.\ dLOP-GP and gPoE-GP, truly support variation in hyperparameters or kernels.

\vspace*{-0.3cm}
\section{Conclusion}
\vspace*{-0.3cm}
We presented a theoretical basis for the gPoE-GP model which was previously believed to be theoretically unfounded but yields surprisingly good results in practice. Intuitions from the new framework, along with a new model, dLOP-GP, that is an improvement over gPoE-GP are validated empirically on three datasets.



\end{document}